\documentclass{article}
\usepackage[preprint]{neurips_2026}
\usepackage[utf8]{inputenc} 
\usepackage[T1]{fontenc}    
\usepackage{hyperref}       
\usepackage{url}            
\usepackage{booktabs}       
\usepackage{amsfonts}       
\usepackage{nicefrac}       
\usepackage{microtype}      
\usepackage{xcolor}         
\usepackage{graphicx}
 \usepackage{times}
\usepackage{latexsym}
\usepackage[T1]{fontenc}
\usepackage[utf8]{inputenc}
\usepackage{microtype}
\usepackage{inconsolata}
\usepackage{graphicx}
\usepackage{booktabs}
\usepackage{amsmath}
\usepackage{multirow}
\usepackage{xcolor}
\usepackage{hyperref}
\usepackage{url}
\usepackage{listings}

\definecolor{darkblue}{rgb}{0.0,0.0,0.55}
\hypersetup{colorlinks=true,linkcolor=darkblue,citecolor=darkblue,urlcolor=darkblue}

\title{Output Language Confusion under Multilingual Prompt Contamination}

\author{
  \textbf{Riju Marwah$^{1,2}$, Ritvik Garimella$^{2}$, Khusham Bansal$^{2,3}$, Atishay Jain$^{2,4}$, Amit Sheth$^{2,5}$} \\
  $^1$University of Tübingen, Germany \\
  $^2$Artificial Intelligence Institute, University of South Carolina, USA \\
  $^3$Thapar Institute of Engineering \& Technology, India\\
  $^4$Indian Institute of Technology Kanpur, India\\
  $^5$Indian AI Research Organization, India \\
  \texttt{riju.marwah@student.uni-tuebingen.de}
}

\begin{document}

\maketitle

\begin{abstract}
Standard factual benchmarks assume clean monolingual prompts and exact-match
scoring, two assumptions that break simultaneously in real-world multilingual deployment, from retrieval-augmented generation pipelines returning mixed-language passages to users pasting multilingual web content. We introduce \textbf{Multilingual Distractor Interference (MDI)}, a lightweight and fully replicable evaluation protocol requiring no new data or annotation, in which factual questions are preceded by a semantically irrelevant foreign-language sentence, and evaluate five instruction-tuned LLMs across TruthfulQA and TriviaQA under eight distractor conditions (40,000 evaluations). Our central finding is a metric confound: for Llama-3.1-8B under a Hindi distractor, 58\% of responses switch to Devanagari script, yielding a raw hallucination proxy of 0.710, but manual review reveals that 120 of 148 script-switched responses that were correct under clean conditions remain semantically correct despite being written in the wrong script, reducing the adjusted semantic hallucination rate to 0.470. All other models respond through abstention escalation with no hallucination increase. A paragraph-length English distractor triggers near-universal abstention (0.806-0.998) across all models, consistent with reading-comprehension confusion, a failure mode with direct consequences for multilingual RAG pipelines. TruthfulQA multiple-choice accuracy is unaffected under all single-sentence conditions. These results show that exact-match hallucination rates in mixed-language settings should be decomposed into script-switching and semantic error components before drawing conclusions about model reliability.
\end{abstract}

\section{Introduction}
\label{sec:intro}

Factual QA benchmarks are a primary tool for measuring LLM reliability, yet two of their core assumptions routinely fail in deployment. First, they assume monolingual prompts \citep{hendrycks2020mmlu, shi2022language}: retrieval-augmented generation (RAG) pipelines frequently return mixed-language passages \citep{asai2023selfrag}, users paste multilingual web content, and cross-lingual queries are common. Second, they assume exact-match scoring accurately reflects semantic error: a model that answers correctly in the wrong script will score identically to one that hallucinated.

These two assumptions interact. When a foreign-language distractor triggers a model to respond in that language, exact-match scoring treats every such response as a hallucination, even when the answer is factually correct. The resulting metric inflation is not uniform across models; it is concentrated in models that have the generative capacity to produce fluent text in the distractor language.
The practical consequence is that clean-prompt benchmarks not only \emph{underestimate} real-world failure for high-capacity multilingual models, but also \emph{mischaracterize} the nature of that failure.

We study this interaction directly. Our setup is not adversarial in the traditional sense: the distractor sentence explicitly instructs the model to ignore it, and the target question is unchanged in English. Any response degradation therefore reflects the model's inability to maintain language-consistent instruction following when the prompt context contains unexpected script boundaries, a realistic failure mode in RAG and document-processing pipelines. To specifically isolate the script-boundary mechanism, we include Spanish as a control condition: sharing the Latin script with English, it separates script-boundary interference from language-identity interference. 

Strikingly, the model most vulnerable to this failure is not the weakest multilingual model but the strongest. This is consistent with what has been termed the \textit{multilingual curse}, where richer multilingual pretraining can introduce cross-lingual interference that degrades output language control rather than improving it. 


Our contributions are three-fold:
\begin{enumerate}
  \item We introduce \textbf{Multilingual Distractor Interference (MDI)}, a
    lightweight, reproducible evaluation protocol for probing factual robustness
    under mixed-language prompt contamination.
  \item We identify and quantify a \textbf{metric confound} in exact-match-based hallucination scoring: for Llama-3.1-8B under Hindi distraction, 120 of 148 script-switched responses that were correct under clean conditions are semantically correct answers written in the wrong script, reducing the raw hallucination proxy from 0.710 to an adjusted semantic hallucination rate of 0.470.
  \item We characterize a \textbf{failure mode taxonomy}: script-switching hallucination, abstention escalation, and stability, showing that the operative mode is determined by model-specific spontaneous generative capacity in the distractor language. The clean dissociation across five models with identical inputs is itself the finding: it rules out general prompt sensitivity and points to a tractable, model-level mechanism.
\end{enumerate}

\begin{figure*}[t]
  \centering
  \includegraphics[width=\textwidth]{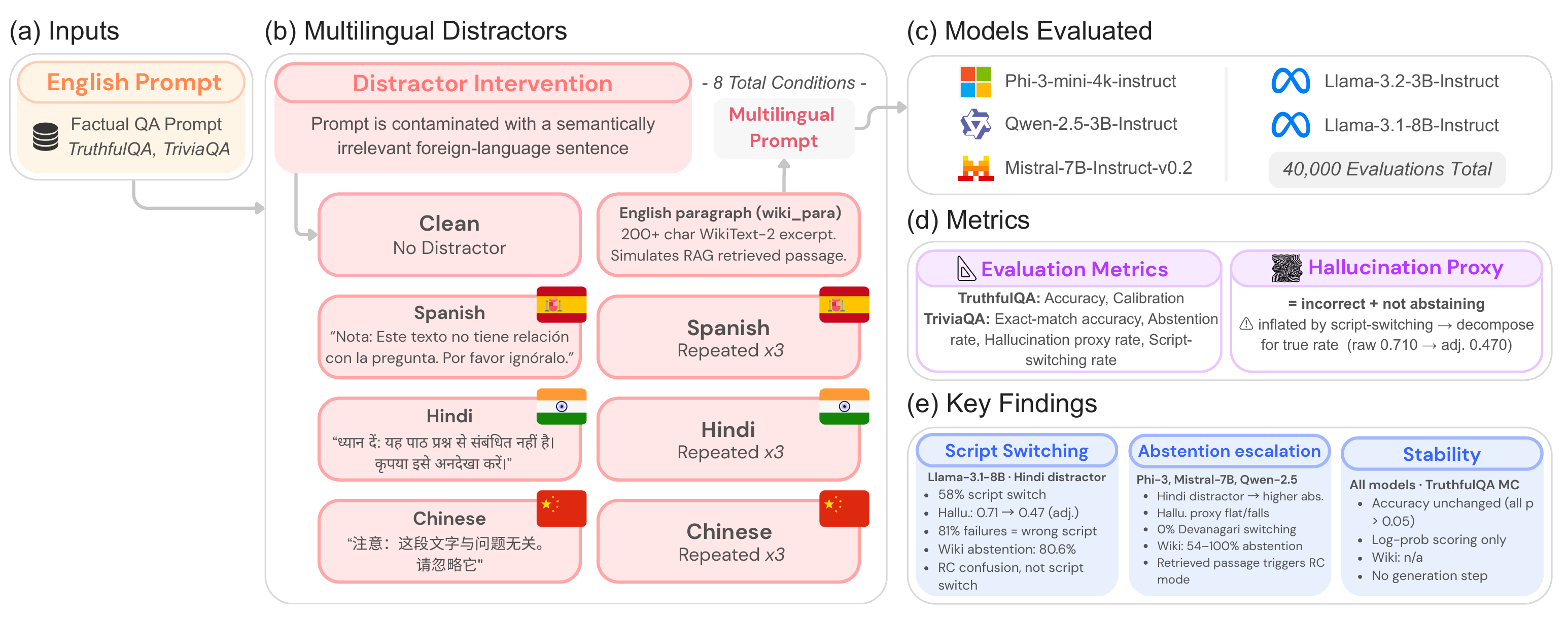}
  \caption{\textbf{Multilingual Distractor Interference (MDI) evaluation pipeline and key findings.}
  \textbf{(a)} An English factual QA prompt is contaminated with a semantically irrelevant foreign-language sentence.
  \textbf{(b)} Eight distractor conditions: clean, Spanish/Hindi/Chinese and their $\times 3$ variants, and a paragraph-length English Wikipedia excerpt (wiki\_para).
  \textbf{(c)} Five instruction-tuned LLMs evaluated under identical deterministic decoding settings.
  \textbf{(d)} Metrics: TruthfulQA accuracy and calibration; TriviaQA exact-match accuracy, abstention rate, hallucination proxy rate, and script-switching rate.
  \textbf{(e)} Three failure modes: script-switching hallucination (Llama-3.1-8B, Hindi), abstention escalation (Phi-3, Mistral-7B, Qwen-2.5), and stability (all models, TruthfulQA MC).
  Raw hallucination proxy of 0.710 adjusts to 0.470 after accounting for script-switched but semantically correct responses identified through manual review.}
  \label{fig:concept}
\end{figure*}

\section{Related Work}
\label{sec:related}

\paragraph{Multilingual evaluation.}
Established multilingual benchmarks such as XTREME~\citep{hu2020xtreme} and XGLUE~\citep{shi2020xglue} evaluate models on tasks presented in a target language, measuring cross-lingual transfer and per-language performance. Open-domain multilingual QA benchmarks such as MKQA~\citep{longpre2021mkqa}
similarly assume that task language and prompt language are aligned. Our setup differs fundamentally: the task remains English throughout; the foreign language appears only as prompt contamination. This distinction matters because it isolates instruction-following robustness from cross-lingual comprehension ability, a dimension that per-language benchmarks do not probe.

\paragraph{Robustness to distracting context.}
\citet{jia2017adversarial} showed that prepending adversarial sentences to reading comprehension passages substantially degraded model accuracy; subsequent work extended this to extractive and generative QA settings. These studies focus on semantically relevant or adversarially crafted distractors within a single language. We study semantically irrelevant distractors that differ from the question in script and language, with an explicit ignore instruction, making any failure a clear case of instruction-following breakdown rather than comprehension difficulty. \citet{muennighoff2023crosslingual} show that instruction-following degrades when instructions and content appear in different languages, which is consistent with our finding that script-boundary crossing in the prompt context disrupts English output generation.

\paragraph{Multilingual contamination in RAG.}
Retrieval-augmented generation pipelines frequently surface passages in languages other than the query language, particularly for cross-lingual or low-resource queries~\citep{asai2023selfrag}. While prior work has examined multilingual retrieval quality and cross-lingual passage selection~\citep{longpre2021mkqa}, systematic study of how mixed-language retrieved passages affect generator output behavior rather than retrieval ranking has been limited. Our \texttt{wiki\_para} ablation (Section~\ref{sec:para}) provides a controlled analog: a monolingual English paragraph prepended to a factual question, isolating the effect of contextual format confusion from script-switching.

\paragraph{Hallucination and factuality metrics.}
Exact-match scoring is known to be an imperfect proxy for semantic correctness in monolingual settings~\citep{joshi2017triviaqa}; we show it introduces a systematic directional bias in multilingual settings. For models with strong spontaneous generation in the distractor language, exact-match conflates script-switched but semantically correct responses with genuine hallucinations, inflating apparent error rates in a way that depends on the model's generative profile rather than its factual accuracy.

\section{Multilingual Distractor Interference (MDI)}
\label{sec:method}

MDI is a lightweight evaluation protocol that probes whether semantically irrelevant foreign-language context disrupts English factual question answering.
The key design choice is that the distractor is not adversarial: it explicitly declares itself irrelevant and instructs the model to ignore it. Any response degradation therefore reflects a failure of language-grounded instruction following rather than a failure of comprehension. The model understood the question but could not maintain English output in the presence of a foreign-script context. This makes MDI a clean probe for a specific and practically relevant robustness dimension. MDI requires no new data collection, annotation, or model access beyond
standard inference, making it straightforwardly applicable to any
instruction-tuned model evaluated on existing factual QA benchmarks.

\subsection{Task Definition}
\label{sec:task}

Each question is evaluated under eight distractor conditions:

\begin{itemize}
  \item \texttt{clean}: no distractor; establishes the per-model baseline.
  \item \texttt{es} / \texttt{es\_long}: a Spanish sentence repeated once or
    three times.
  \item \texttt{hi} / \texttt{hi\_long}: a Hindi sentence in Devanagari script,
    repeated once or three times.
  \item \texttt{zh} / \texttt{zh\_long}: a Chinese sentence in Simplified
    Chinese script, repeated once or three times.
  \item \texttt{wiki\_para}: a paragraph-length English Wikipedia excerpt,
    included as an ablation to isolate context-length and format-confusion
    effects from script-switching.
\end{itemize}

The three distractor languages were selected to vary along two dimensions relevant to the hypothesized failure mechanism: script and pretraining resource level. Spanish uses the Latin script shared with English, providing a baseline condition where script boundaries are absent and any interference must arise from language identity alone. Hindi (Devanagari script) and Chinese (Simplified Chinese script) both introduce a non-Latin script boundary, but differ substantially in pretraining corpus size: Chinese is among the highest-resource non-English languages in most LLM pretraining corpora, while Hindi is comparatively lower-resource. This contrast allows us to test whether the failure mode is driven by script-switching propensity specifically or by distractor language resource level more generally, a distinction that proves consequential in Section~\ref{sec:mechanism}.

The base distractor for each language is a single sentence that declares itself
irrelevant and instructs the model to ignore it. For example, the Spanish distractor reads: \textit{``Nota: Este texto no tiene relaci\'{o}n con la pregunta. Por favor ign\'{o}ralo.''} Hindi and Chinese equivalents are constructed analogously in their respective native scripts. The long variants repeat the base sentence three times to test whether distractor length modulates the observed effects. The \texttt{wiki\_para} condition uses a fixed paragraph sampled from the WikiText-2 training split (minimum 200 characters) as a realistic approximation of a retrieved English passage. In all conditions, the distractor is prepended to the question with a blank line separator.

\subsection{Prompt Format}
\label{sec:prompt}

For TruthfulQA, each item is presented as a four-choice multiple-choice question. The system prompt instructs the model to respond with a single letter
(A-D); accuracy is measured by log-probability scoring over the four option tokens, requiring no generation. This format structurally prevents script-switching and serves as a format-controlled upper bound on robustness.

For TriviaQA, each item is presented as an open-ended short-answer question.
The system prompt instructs the model to respond with a short phrase only, and specifies \textit{``If you do not know, say: I don't know.''} to elicit calibrated abstention. All TriviaQA generation uses greedy decoding ($T\!=\!0$, maximum 32 new tokens).

\subsection{Metrics}
\label{sec:metrics}

\paragraph{TruthfulQA.} Accuracy is reported, defined as the proportion of questions where the highest-probability option is correct.

\paragraph{TriviaQA.} Four metrics are reported: (i)~\emph{exact-match
accuracy}, evaluated against the full gold alias list; (ii)~\emph{abstention rate}, detected by substring match on ``I don't know,'' ``unknown,'' and
related phrases; (iii)~\emph{hallucination proxy rate}, defined as the
proportion of responses that are neither correct nor abstaining, such that accuracy~$+$~abstention~$+$~hallucination proxy~$= 1$; and (iv)~\emph{script-switching rate}, the proportion of outputs containing at least one character from the distractor's native script, detected via Unicode block membership (Devanagari for Hindi; CJK Unified Ideographs for Chinese).

The hallucination proxy is an intentionally conservative operationalization of confident factual error. As we show in Section~\ref{sec:confound}, it
conflates script-switched but semantically correct responses with genuine semantic errors in multilingual settings, and must be decomposed accordingly.

\paragraph{Statistical testing.} Pairwise comparisons between each distractor condition and the clean baseline use McNemar's test on paired binary outcomes with continuity correction, with 95\% confidence intervals computed via bootstrap resampling ($n\!=\!2000$).

\section{Experimental Setup}
\label{sec:setup}

\subsection{Models}

We evaluate five publicly available instruction-tuned models spanning four
training lineages and two parameter scales (Table~\ref{tab:models}). Model
selection was guided by three criteria. First, \emph{family diversity}: Meta
Llama, Microsoft Phi, Mistral AI, and Alibaba Qwen represent independent
pretraining pipelines with different multilingual data compositions, allowing
failure mode differences to be attributed to training rather than shared
architectural choices. Second, \emph{scale coverage}: the 3-3.8B and 7-8B
ranges are representative of widely deployed open-source models that fit within
single-GPU inference budgets in float16 precision, enabling exact replication
without proprietary access. Third, \emph{controlled intra-family comparison}:
Llama-3.1-8B and Llama-3.2-3B share the same tokenizer, so any behavioral
difference between them under identical distractor conditions cannot be
attributed to tokenizer fragmentation, isolating model capacity and pretraining
exposure as the operative variables.

\begin{table}[h]
\centering
\small
\caption{Models evaluated. All loaded in float16 on a single GPU via
  HuggingFace Transformers; inference uses greedy decoding throughout.
  Language support as stated in official model cards.}
\label{tab:models}
\resizebox{\columnwidth}{!}{%
\begin{tabular}{llrl}
\toprule
Model & Family & Params & Supported languages \\
\midrule
Llama-3.1-8B-Instruct   & Meta Llama    & 8B   & 8 (incl.\ Hindi, Chinese$^\dagger$) \\
Llama-3.2-3B-Instruct   & Meta Llama    & 3B   & 8 (incl.\ Hindi, Chinese$^\dagger$) \\
Phi-3-mini-4k-instruct  & Microsoft Phi & 3.8B & English-primary \\
Mistral-7B-Instruct-v0.2 & Mistral AI   & 7B   & English-primary \\
Qwen2.5-3B-Instruct     & Alibaba Qwen  & 3B   & 29+ (incl.\ Chinese) \\
\bottomrule
\end{tabular}}
\vspace{2pt}
\footnotesize{$^\dagger$Llama 3.x officially supports Hindi but not Chinese;
both models were trained on a broader set of languages than the 8 officially
supported.}
\end{table}

\subsection{Datasets}

We evaluate on TruthfulQA~\citep{lin2022truthfulqa} and TriviaQA~\citep{joshi2017triviaqa}, chosen for their complementary evaluation formats. TruthfulQA's multiple-choice structure permits log-probability scoring over single option tokens without any generation step, providing a format-controlled baseline in which script-switching is structurally impossible. TriviaQA's open short-answer format exposes generation behavior directly, allowing observation of script-switching, abstention, and hallucination under distractor conditions. Together, the two datasets allow us to separate failure modes that arise from the generation process from those that arise from answer selection, a distinction that is central to interpreting our results.

For TruthfulQA we use the mc1 split from the validation set, sampling 500
questions with a fixed random seed (seed\,=\,42). For TriviaQA we use the \texttt{unfiltered.nocontext} configuration from the validation split, sampling
500 questions with the same seed; answers are evaluated against the full alias list provided in the dataset.

\subsection{Evaluation Scale}

Each model is evaluated on every question under every distractor condition,
yielding:
$5 \text{ models} \times 2 \text{ datasets} \times 8 \text{ conditions} \times 500 \text{ questions} = 40{,}000 \text{ evaluations.}$

Question-distractor pairings are fixed across models so that all comparisons
are paired, enabling McNemar's test for significance testing throughout.

\section{Results}
\label{sec:results}

\subsection{TruthfulQA: Multiple-Choice Is Robust}
\label{sec:tqa}

TruthfulQA accuracy is unaffected by single-sentence multilingual distractors
for all models (Table~\ref{tab:tqa}). No accuracy drop is statistically
significant under single-sentence conditions ($p > 0.05$, McNemar's test).

\begin{table}[h]
\centering
\small
\caption{TruthfulQA accuracy: clean baseline and worst distractor condition.
  95\% CIs via bootstrap ($n{=}2000$). \emph{ns}: not significant;
  ${}^{**}p{<}0.01$; ${}^{***}p{<}0.001$.}
\label{tab:tqa}
\begin{tabular}{lllr@{\hspace{4pt}}l}
\toprule
Model & Clean & Worst & $\Delta$ & $p$ \\
\midrule
Llama-3.1-8B & 0.510 & 0.478 (\texttt{wiki}) & $-$0.032 & \emph{ns} \\
Llama-3.2-3B & 0.554 & 0.454 (\texttt{zh\_l}) & $-$0.100 & ${}^{***}$ \\
Phi-3-mini   & 0.652 & 0.616 (\texttt{wiki}) & $-$0.036 & \emph{ns} \\
Mistral-7B   & 0.678 & 0.626 (\texttt{wiki}) & $-$0.052 & ${}^{**}$ \\
Qwen-2.5-3B  & 0.440 & 0.440 (\texttt{es\_l}) & $+$0.000 & \emph{ns} \\
\bottomrule
\end{tabular}
\end{table}

The insulation of TruthfulQA from single-sentence distractors follows directly from its format: log-probability scoring over single letter tokens removes the generation step that makes script-switching possible. This makes TruthfulQA MC accuracy an unreliable signal in either direction under distractor conditions, a methodological point for future benchmark design.

Two observations in Table~\ref{tab:tqa} warrant brief note. First, Phi-3-mini
and Qwen-2.5-3B show small \emph{positive} accuracy deltas under Hindi and Chinese distractors (e.g., Phi-3-mini reaches 0.770 under Hindi, $\Delta=+0.118$, $p<0.001$). We do not interpret this as a robustness benefit; log-probability scoring over single tokens (A/B/C/D) is sensitive to any shift in the model's token distribution, and foreign-language context can shift that distribution in ways that happen to favor the correct letter by chance. This artefact reinforces the point that TruthfulQA MC accuracy is not a reliable signal in either direction under distractor conditions.

Second, Llama-3.2-3B shows a consistent and significant accuracy drop under most distractor conditions, including a $\Delta=-0.100$ drop under \texttt{zh\_long} ($p<0.001$), the largest absolute drop in Table~\ref{tab:tqa}. Unlike the script-switching mechanism in Llama-3.1-8B, this effect occurs on a multiple-choice task where generation is not involved, suggesting a different mechanism: the longer Chinese distractor may shift the log-probability mass over option tokens in a direction that consistently disfavors the correct letter for this model. We flag this as a distinct sensitivity worth investigating but do not pursue it further here, as the effect is small in absolute terms and absent from the open-answer results.

\subsection{TriviaQA: A Clean Failure Mode Dissociation}
\label{sec:trivia}

TriviaQA open-answer generation reveals large, model-specific effects. The Hindi distractor produces the most divergent cross-model behavior (Table~\ref{tab:trivia}).

\begin{table}[h]
\centering
\small
\setlength{\tabcolsep}{3.5pt}
\caption{TriviaQA results under Hindi distractor. ``Deva\%'' = Devanagari
script-switching rate. $\Delta$ columns are distractor $-$ clean.
${}^{***}p{<}0.001$; \emph{ns}: not significant.}
\label{tab:trivia}

\begin{tabular}{lccccc}
\toprule
& \multicolumn{2}{c}{Accuracy} & \multicolumn{2}{c}{Hallu.\ Proxy} & \\
\cmidrule(lr){2-3}\cmidrule(lr){4-5}
Model & Clean & $\Delta$ & Clean & $\Delta$ & Deva\% \\
\midrule
Llama-3.1-8B & 0.518 & $-0.290^{***}$ & 0.416 & $+0.294^{***}$ & 58.2 \\
Llama-3.2-3B & 0.410 & $-0.052^{***}$ & 0.518 & $+0.018^{\mathrm{ns}}$ & 1.2 \\
Phi-3-mini   & 0.404 & $-0.020^{\mathrm{ns}}$ & 0.514 & $-0.146^{***}$ & 1.0 \\
Mistral-7B   & 0.158 & $+0.010^{\mathrm{ns}}$ & 0.744 & $-0.162^{***}$ & 0.6 \\
Qwen-2.5-3B  & 0.266 & $-0.124^{***}$ & 0.382 & $-0.254^{***}$ & 0.0 \\
\bottomrule
\end{tabular}
\end{table}

\begin{figure*}[t]
\centering
\includegraphics[width=\textwidth]{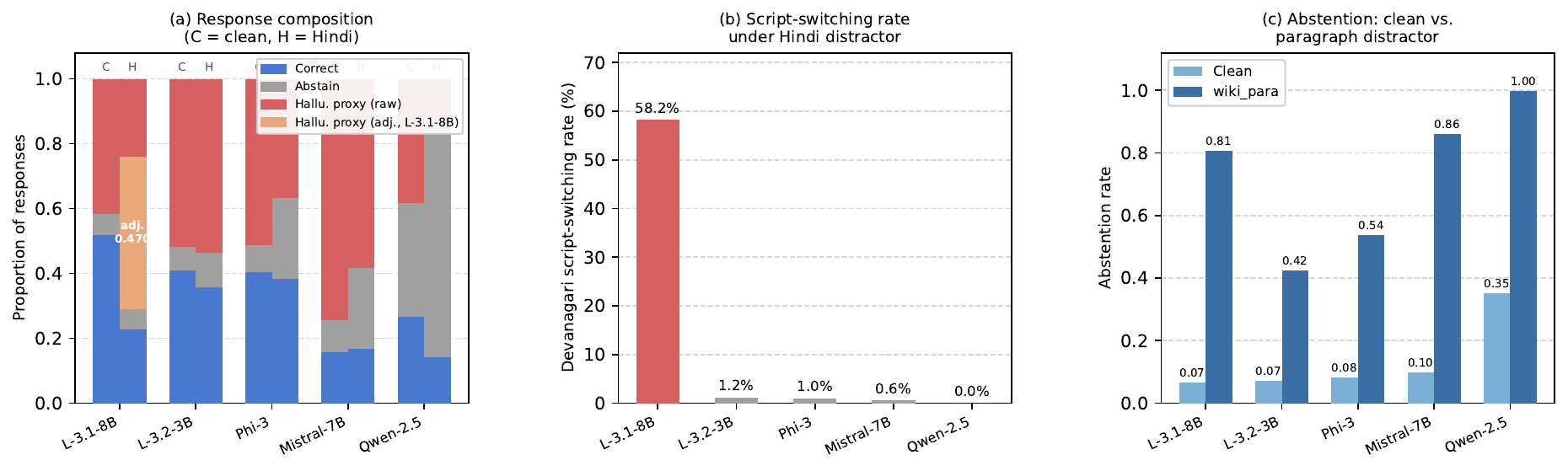}
\caption{
  \textbf{(a)} TriviaQA response composition (correct / abstain / hallucination
  proxy) for each model under clean (C) and Hindi (H) distractor conditions.
  For Llama-3.1-8B, the orange segment shows the adjusted semantic hallucination
  rate (0.470) after removing script-switched but semantically correct responses
  from the raw proxy (0.710).
  \textbf{(b)} Devanagari script-switching rate under the Hindi distractor.
  Only Llama-3.1-8B switches spontaneously at scale.
  \textbf{(c)} Abstention rate under clean vs.\ paragraph-length English
  distractor (\texttt{wiki\_para}), showing near-universal abstention across
  all models.
}
\label{fig:main}
\end{figure*}

The dissociation is striking (Figure~\ref{fig:main}). Llama-3.1-8B switches to Devanagari on 58.2\% of
responses; 75.2\% of those script-switched outputs fail exact-match evaluation. Accuracy drops 29 pp ($p<0.001$) while the hallucination proxy rises 29.4 pp ($p<0.001$) and abstention rate is essentially unchanged (0.066 $\to$ 0.062): the model is answering confidently, in the wrong language.

All other models produce Devanagari on fewer than 1.2\% of responses.
Critically, their hallucination rates do not rise; for Phi-3-mini, Mistral-7B, and Qwen-2.5-3B, hallucination rates fall significantly as abstention increases. These models are pulled toward refusal, not toward language-confused generation.

Abstention rates under the Hindi distractor and the \texttt{wiki\_para} condition
are shown in Table~\ref{tab:abstain}.

\begin{table}[h]
\centering
\small
\caption{TriviaQA abstention rates under Hindi and \texttt{wiki\_para} conditions.}
\label{tab:abstain}
\begin{tabular}{l r r r r}
\toprule
Model & Clean & Hindi & $\Delta$Hi & \texttt{wiki} \\
\midrule
Llama-3.1-8B & 0.066 & 0.062 & $-$0.004 & 0.806 \\
Llama-3.2-3B & 0.072 & 0.106 & $+$0.034 & 0.424 \\
Phi-3-mini   & 0.082 & 0.248 & $+$0.166 & 0.538 \\
Mistral-7B   & 0.098 & 0.250 & $+$0.152 & 0.860 \\
Qwen-2.5-3B  & 0.352 & 0.730 & $+$0.378 & 0.998 \\
\bottomrule
\end{tabular}
\end{table}

\subsection{The Metric Confound: Unpacking the Hallucination Proxy}
\label{sec:confound}

The hallucination proxy of 0.710 for Llama-3.1-8B under Hindi distraction is the paper's most alarming headline number, and also the most misleading. Of the 291 script-switched responses, we reviewed all 291 cases for semantic correctness. Among these, 148 responses were correct under the clean condition but script-switched under the Hindi distractor. Of these 148 responses, 120 (81.1\%) were semantically correct answers written in Devanagari script, while the remaining 28 (18.9\%) were genuine semantic errors. For example, ``Colombia.'' rendered as its Devanagari transliteration, or ``Oregon.'' similarly transcribed.

The remaining 28 cases within this 148-response subset (18.9\%) are genuine semantic errors: wrong proper nouns or semantically incorrect paraphrases. Accounting for the 120 script-switched responses identified as semantically correct, the adjusted semantic hallucination rate is $235/500 = 0.470$, compared to a raw hallucination proxy of 0.710 and a clean baseline of 0.416. The primary consequence of script-switching is \emph{output in the wrong language}, not fabricated content, though a meaningful real increase in semantic errors (+5.4 pp) remains.

\textbf{The implication for evaluation practice is direct}: exact-match-based hallucination metrics in multilingual settings should be decomposed into script-switched (potentially semantically correct) and genuinely wrong components. Reporting raw exact-match failure for a model like Llama-3.1-8B in mixed-language conditions overstates semantic hallucination by a factor of
approximately 1.5$\times$. Benchmark comparisons between multilingual and English-primary models may hence systematically understate the reliability gap in contexts where the multilingual model is more likely to script-switch.

\subsection{Script-Switching as the Proximal Mechanism}
\label{sec:mechanism}

Two comparisons rule out alternative explanations for the Llama-3.1-8B
dissociation.

\textbf{Hindi vs.\ Chinese.} Chinese is a higher-resource language than Hindi in most LLM pretraining corpora. If distractor language resource level drove the effect, Llama-3.1-8B should show an equal or stronger hallucination spike under the Chinese distractor. It does not: CJK-script outputs occur on 0\% of responses, and the hallucination rate falls slightly to 0.382 ($\Delta=-0.034$, $p=0.009$). The model has sufficient Hindi generative capacity to switch into Devanagari output spontaneously, but not Chinese.

\textbf{Llama-3.1-8B vs.\ Llama-3.2-3B.} Both models share the same tokenizer, ruling out tokenizer fragmentation as the driver. Yet Llama-3.2-3B produces Devanagari on only 1.2\% of responses under the same Hindi distractor. The relevant variable is not tokenization, and not instructed generation ability (both models can produce Devanagari when explicitly asked to do so), but spontaneous script-switching propensity under distractor conditions. This propensity appears to be a function of model scale and pretraining Hindi exposure, and it is concentrated in the larger model, not the smaller one.

This last point is counterintuitive and practically important:
\textbf{stronger multilingual capacity does not protect against this failure mode; for models with strong spontaneous generation in the distractor language, it produces it.} This is consistent with what has been termed the \emph{multilingual curse}, the observation that incorporating more languages during pretraining can introduce cross-lingual interference that degrades target-language behavior~\citep{muennighoff2023crosslingual}. Here the interference manifests not as degraded comprehension but as degraded output language control: the model understands the English question but cannot suppress generation in the language whose script dominates the prompt context. The implication is that models with the richest multilingual pretraining may be the most susceptible to this specific failure mode, not the least.

\subsection{Paragraph-Length Distractor}
\label{sec:para}

The \texttt{wiki\_para} condition produces a qualitatively different and cross-model consistent effect: near-universal abstention (Mistral-7B: 0.860; Qwen-2.5-3B: 0.998; even Llama-3.1-8B, which shows minimal abstention under Hindi, abstains on 80.6\% of \texttt{wiki\_para} questions). This is not a script-switching effect as the paragraph is English, but a context-format confusion: models appear to treat the prepended paragraph as part of the question context and decline to answer when the ``passage'' does not contain the answer, as if performing reading comprehension rather than open-domain QA.

This is directly relevant to RAG deployments, and to a gap in how RAG systems are currently evaluated. Most existing evaluation of RAG systems focuses on retrieval quality and answer accuracy over clean, language-matched passages; systematic evaluation over mixed-language or contextually mismatched retrieved context has been limited. Our results suggest that such mismatches can cause widespread abstention independent of retrieval quality. A model that retrieves the correct passage but treats it as reading comprehension context will abstain rather than answer. This is a failure mode that is invisible to retrieval-focused evaluation metrics but consequential in deployment. Evaluation protocols for multilingual RAG should therefore include conditions where retrieved passages are in a different language or format than the query, and should track abstention rates alongside answer accuracy.

\subsection{Distractor Length Effects}
\label{sec:length}

Repeating the single-sentence distractor three times produces modest additional degradation but does not change the qualitative pattern. For Llama-3.1-8B, \texttt{hi\_long} hallucination proxy is 0.704 vs.\ 0.710 for \texttt{hi}, essentially unchanged. For Qwen-2.5-3B, abstention rises from 0.730 to 0.842 under \texttt{hi\_long}. The failure mode taxonomy established under single-sentence conditions is stable under distractor repetition.

\section{Discussion}
\label{sec:discussion}

\paragraph{Why does Llama-3.1-8B switch to Devanagari?}
The pattern is consistent with a model that has learned strong Hindi text generation during pretraining and whose instruction-following mechanism is insufficient to suppress spontaneous language switching when a Devanagari-script context is present. Crucially, the relevant capacity is \emph{spontaneous} rather than instructed: all five models in our evaluation can produce Devanagari output when explicitly instructed to do so, but only Llama-3.1-8B switches scripts without such instruction under distractor conditions. The 8B/3B contrast within the Llama family, with identical tokenizers but different scale and pretraining data exposure, suggests that the failure mode scales with multilingual capacity in a way that is not obvious from standard multilingual benchmarks. We do not have access to training data composition to make this definitive.

\paragraph{A failure mode taxonomy.}
Our results support three response modes under multilingual distractor text: (1)~\emph{script-switching hallucination}: confident generation in the wrong language, inflating exact-match error; (2)~\emph{abstention escalation}: rising ``I don't know'' rates with no hallucination increase; and (3)~\emph{stability}: no significant change in either direction. Which mode a model exhibits in open-answer settings depends on whether it has spontaneous generative capacity in the distractor language, not on distractor length, language resource level, or tokenizer characteristics. A fourth route to apparent stability is format constraint: TruthfulQA's multiple-choice scoring removes the generation step entirely, suppressing the failure mode regardless of the model's multilingual profile.

\paragraph{Implications for evaluation.}
Current multilingual robustness evaluations focus on per-language task performance. Our results suggest that a different dimension is both practically important and poorly captured by existing protocols: the effect of foreign-language prompt context on otherwise English tasks. More concretely: (i) hallucination proxy rates computed via exact-match in mixed-language settings should be reported alongside script-switching rates and adjusted semantic error rates; (ii) models explicitly optimized for multilingual instruction following should be evaluated specifically for output language control under mixed-script prompt conditions, as stronger multilingual capacity appears to increase rather than decrease susceptibility to this failure mode; (iii)~RAG evaluation
benchmarks should include mixed-language retrieved context, as even
English-language paragraph noise produces widespread abstention. More broadly, these findings suggest that multilingual pretraining objectives and data composition have consequences for output language control that are not captured by standard cross-lingual transfer benchmarks: a dimension that warrants
explicit attention in multilingual model development.

\paragraph{Limitations.}
Our experimental choices reflect deliberate scope decisions appropriate for a diagnostic probe. All models are evaluated at realistic single-GPU inference budgets in float16 precision, reflecting conditions under which these models are actually deployed rather than idealized research settings. The distractor sentences explicitly declare themselves irrelevant by design: this isolates instruction-following failure from comprehension failure, making the probe interpretable rather than weaker. More naturalistic contamination without an ignore instruction is important future work and may produce larger effects. We evaluate five open-source models; the pattern may not generalize to closed-source models or models with different pretraining data mixes, and we encourage replication of MDI across a broader model population. Our model set consists of models released between 2023 and 2024 and therefore does not include the latest generations of open-source instruction-tuned LLMs. We view replication across current-generation models as an important direction for future work. Abstention detection via substring matching may miss non-standard refusal formulations, particularly Mistral-7B's more discursive refusals. 


\section{Conclusion}
\label{sec:conclusion}

We introduce Multilingual Distractor Interference (MDI) and use it to identify a systematic metric confound in exact-match-based hallucination scoring: for models with strong spontaneous Hindi generative capacity, a Hindi-language distractor triggers script-switching rather than semantic hallucination, and exact-match scoring cannot distinguish the two. Review of all 291 script-switched responses finds that, among the 148 cases that were correct under clean conditions, 120 (81.1\%) are semantically correct answers in the wrong script, reducing the raw hallucination proxy from 0.710 to an adjusted semantic hallucination rate of 0.470. Models without this spontaneous capacity respond through abstention escalation instead, a less severe but equally systematic failure mode with different practical consequences. Paragraph-length English distractors produce a third mode: near-universal abstention consistent with reading-comprehension confusion,
with direct implications for RAG evaluation. The takeaway for practitioners is two-fold: (1) deploy-time failure mode in mixed-language pipelines depends on the model's spontaneous generative profile; and (2) exact-match hallucination metrics reported for models with strong non-English generation should be treated as upper bounds requiring script-switch decomposition before interpretation. More broadly, our results suggest that output language control under mixed-script prompt conditions is a distinct and underexamined axis of multilingual model quality, and one that should be evaluated explicitly alongside cross-lingual transfer and per-language accuracy as multilingual models continue to scale.





\bibliographystyle{plainnat}
\bibliography{custom}

\end{document}